# Data Generation for Post-OCR correction of Cyrillic handwriting

Evgenii Davydkin, Aleksandr Markelov, Egor Iuldashev, Anton Dudkin, Ivan Krivorotov
*Dbrain*

ed@dbrain.io, am@dbrain.io, ey@dbrain.io, addudkin@dbrain.io, ik@dbrain.io

This paper introduces a novel approach to post-Optical Character Recognition Correction (POC) for handwritten Cyrillic text, addressing a significant gap in current research methodologies. This gap is due to the lack of large text corporas that provide OCR errors for further training of language-based POC models, which are demanding in terms of corpora size. Our study primarily focuses on the development and application of a synthetic handwriting generation engine based on Bézier curves. Such an engine generates highly realistic handwritten text in any amounts, which we utilize to create a substantial dataset by transforming Russian text corpora sourced from the internet. We apply a Handwritten Text Recognition (HTR) model to this dataset to identify OCR errors, forming the basis for our POC model training. The correction model is trained on a 90-symbol input context, utilizing a pre-trained T5 architecture with a seq2seq correction task. We evaluate our approach on HWR200 and School_notebooks_RU datasets as they provide significant challenges in the HTR domain. Furthermore, POC can be used to highlight errors for teachers, evaluating student performance. This can be done simply by comparing sentences before and after correction, displaying differences in text. Our primary contribution lies in the innovative use of Bézier curves for Cyrillic text generation and subsequent error correction using a specialized POC model. We validate our approach by presenting Word Accuracy Rate (WAR) and Character Accuracy Rate (CAR) results, both with and without POC correction, using real open corporas of handwritten Cyrillic text. These results, coupled with our methodology, are designed to be reproducible, paving the way for further advancements in the field of OCR and handwritten text analysis. Paper contributions can be found in https://github.com/dbrainio/CyrillicHandwritingPOC

## INTRODUCTION

In the evolving landscape of digitization and text processing, Optical Character Recognition (OCR) has emerged as a pivotal technology, transforming physical texts into digital formats. This transformation has been particularly crucial for handwritten texts, where OCR's role extends beyond digitization to enabling text searchability and analysis. Despite significant advancements in OCR technology, especially in the context of widely-used scripts like Latin, challenges persist in accurately recognizing and converting less commonly digitized scripts, such as Cyrillic handwriting.

The accurate digitization of Cyrillic handwritten text is not only a technological challenge but also a cultural imperative. It enables the preservation and accessibility of a vast array of historical[1] and contemporary documents[2], thus facilitating research and cultural understanding. However, the inherent irregularities and idiosyncrasies of handwritten text present unique challenges for OCR, often resulting in errors that can significantly impact the utility of the digitized content. These errors, typically manifested as insertions, deletions, substitutions, and misinterpretation of characters, necessitate effective post-OCR correction mechanisms.

To address these challenges, our research introduces an innovative approach, leveraging the capabilities of deep learning and synthetic handwriting generation. We employ the T5 transformer model[3] in the seq2seq task of Post-OCR correction (POC). The cornerstone of our method is the generation of synthetic handwritten Cyrillic text using Bézier curves[4]. This technique, novel in the realm of Cyrillic OCR, enables us to create a realistic and extensive dataset derived from Russian text corpora.

Our method involves generating handwritten text from these corpora, followed by its recognition using a specialized Handwritten Text Recognition (HTR) model to identify errors. The subsequent training of our Post-OCR correction model, focusing on a 90-symbol context, is a pivotal step in refining the digitization process. Our research presents a groundbreaking approach in the field of OCR and handwritten text analysis, particularly for the Cyrillic script. We demonstrate the efficacy of our method through comprehensive results, including Word Accuracy Rate (WAR) and Character Accuracy Rate (CAR), with and without Post-OCR Correction on real open corpora of handwritten Cyrillic text. These results not only validate our approach but also provide a reproducible framework for future advancements in this domain.

The novelty of our research lies in its unique combination of synthetic handwriting generation using Bézier curves and the application of deep learning models for post-OCR error correction, an approach previously unexplored for Cyrillic text. Through this study, we aim to bridge the gap in OCR technology for Cyrillic script, enhancing the accuracy and reliability of digitizing handwritten texts.

We evaluate our approach using two open datasets of student essays: HWR200[5] and School_notebooks_RU[6]. Additional evaluations are conducted on our in-house dataset of student essays for practicing essay writing. Our focus on student handwriting is two-fold:

1. Underdeveloped handwriting of students provides additional challenge in HTR task, thus increasing value of POC model in terms of improving readability of recognized text

2. Trained POC model can be later used as a highlighting tool for teachers to focus on students' mistakes and cases of hard-to-recognize handwriting, improving feedback loop.

Our contributions are following:

- We demonstrate a novel approach in generating realistic looking handwritten text which may be used not only in POC tasks but also for handwritten model pretraining. This approach may be applied to other languages, provided that a template letters database is gathered.

- We demonstrate that POC models, trained on OCR errors gathered from predictions on synthetic data, are capable of significant improvement of WAR on student handwriting datasets.

- We provide access to gathered corpora of student essays for domain training of essay-related text tasks

- We provide access to a dataset sample of generated Russian handwritten essays, which can be used not only for POC models training but also for HTR pretraining.

## METHODOLOGY

## 1.1. Handwriting generation approaches overview

### 1.1.1. Limitations of Handwritten-like Fonts for OCR Data Collection

In the pursuit of enhancing OCR error detection and correction capabilities, the utilization of handwritten-like fonts has been commonly employed[7]. However, this strategy exhibits significant limitations in the context of training robust OCR models. The primary drawback stems from the uniformity of these fonts; each occurrence of a letter is essentially a pixel-by-pixel duplicate of its counterparts. This homogeneity leads to a high risk of model overfitting, where the OCR model becomes adept at recognizing these specific font patterns but fails to generalize effectively to real-world handwriting variations. Consequently, an OCR model trained on such data shows limited practical utility when confronted with naturally varying handwritten text.

### 1.1.2. Challenges in Utilizing Keypoint Handwriting Datasets

An alternative methodology involves training a network on keypoint handwriting datasets[8]. While this approach offers a more realistic representation of handwriting variability, it is beset with logistical and technical hurdles. The acquisition of such datasets necessitates specialized hardware and extensive participation from diverse individuals, making the data collection process both time-consuming and expensive. Additionally, training a generative model on this data and deploying it for inference is resource-intensive, imposing significant computational costs and potentially limiting scalability.

### 1.1.3. Constraints in Gathering Real Handwritten Data

Acquiring real handwritten data presents its own set of challenges. For the Post-OCR Correction (POC) model to effectively leverage language patterns, it requires a large and diverse corpus of handwritten text. The creation of such a corpus is a labor-intensive and costly endeavor, impractical for the scope of most research projects. Furthermore, the variability in real handwriting introduces complexities in standardization and digitization, hindering the consistent training and evaluation of OCR models.

### 1.1.4. Proposed Solution: Parameterized Synthetic Handwriting Generation

In response to these challenges, we propose a novel solution for generating synthetic handwriting data, optimally tailored for training OCR models. Our approach hinges on the creation of a base set of handwritten Cyrillic characters. From this foundational dataset, we employ a parametric modification and randomization technique to generate a diverse array of handwriting styles. This method is grounded in the use of Bézier curves to simulate the natural variations and fluidity found in human handwriting.

The advantage of this approach is manifold. Firstly, it circumvents the need for large-scale data collection, thereby reducing time and financial costs. Secondly, it offers a controlled environment to introduce a wide range of handwriting variations, enhancing the robustness and generalizability of the trained OCR model. Finally, the use of Bézier curves allows for a realistic representation of handwriting, closely mirroring the nuances and idiosyncrasies inherent in human-written text.

Visual comparison of all presented approaches is shown on fig. 1.

fig. 1 - Visual comparison of handwritten Cyrillic text

a. Handwritten-like Font

b. Keypoint-generated handwriting

c. Real handwriting

d. Bézier curve handwriting(ours)

In summary, our methodology presents a scalable, cost-effective, and realistic approach to generating synthetic handwritten Cyrillic text, addressing the key limitations of existing data collection methods for OCR training and providing a solid foundation for the subsequent development of an efficient Post-OCR correction model.

## 1.2. Synthetic handwriting generation

### 1.2.1. Bézier curves

In our methodology, Bézier curves serve as the foundational mathematical model for generating synthetic handwritten Cyrillic text. A Bézier curve is a parametric curve extensively employed in computer graphics, defined by a set of control points that dictate its shape. These curves are particularly adept at approximating complex shapes and are scalable, making them ideal for replicating the nuances of handwriting.

We specifically employ four-point Bézier curves, denoted by control points P1, P2, P3, and P4(see fig. 2). These points yield two vectors, which collectively define the curvature of a segment. By manipulating these control points, we can generate a wide variety of curves, each mimicking different handwriting styles. This flexibility is crucial for our application, as it allows for the creation of diverse and realistic handwritten characters. The curve itself can be described as:

$$P = (1 - t)^3 P1 + 3(1 - t)^2 P2 + 3(1 - t)t^2 P3 + t^3 P4 \ (1)$$

fig. 2 - Bézier curve visualization

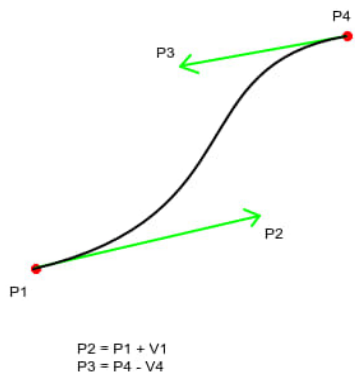

$$P2 = P1 + V1$$
$$P3 = P4 - V4$$

The construction of a complete handwritten letter involves the concatenation of several such Bézier curve segments. To ensure a natural and continuous flow in the handwritten letters, we align the connection points of these segments(see fig. 3). This is achieved by adjusting the vectors at these points to be equal in length and opposite in direction, facilitating a smooth transition from one curve to the next.

fig. 3 - Connection points alignment

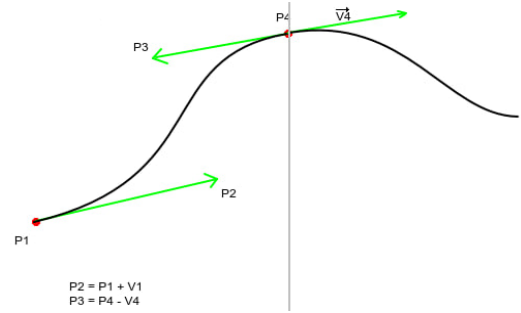

$$P2 = P1 + V1$$
$$P3 = P4 - V4$$

The utilization of Bézier curves in this manner allows us to precisely control the shape of each letter, thereby closely approximating the variability inherent in human handwriting. This approach is not only computationally efficient but also provides a high degree of control over the generated handwriting styles, significantly enhancing the authenticity of the synthetic handwritten dataset used for OCR training and error correction.

### 1.2.2 Character Bézier templates

Our generated handwriting consists of consecutive letter templates connected to each other. They in turn are composed of a series of Bézier curves, each defined by strategically placed control points denoted by circles(see fig. 4) and labeled as points **'Pi'**. These control points are the anchors from which the curves emanate and through which they are manipulated. Adjacent to each control point is a corresponding rectangle, which represents the terminal point of a vector originating from the control point. These vectors are denoted as **'Vi'**.

fig. 4 - Template letter 'IO' of a Russian alphabet

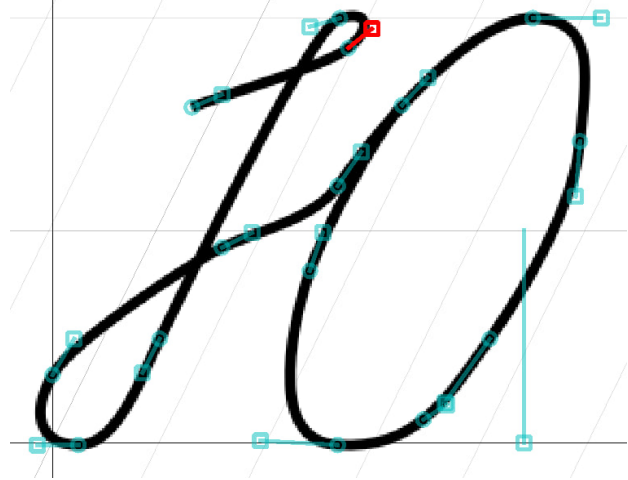

Each vector **'Vi'** associated with a control point 'Pi' is crucial for shaping the curve, as it determines the direction and magnitude of the curve's tangency at that point. The vectors are meticulously designed so that each **'Vi'** has a single opposing vector **'-Vi'**, ensuring that the curvature of the letter is smooth and continuous. This precise vector pairing is fundamental for achieving the seamless transitions between the segments of the Bézier curve.

To construct a letterform, these curves are linked together, with the terminal vector of one curve seamlessly transitioning into the initial vector of the subsequent curve. This is facilitated by the positioning of the last control point of one letter and the first control point of the next. The alignment of these points allows for the creation of cursive or connected letterforms, enabling the simulation of handwriting where letters are naturally joined.

The methodology of employing Bézier curves in this manner allows for the creation of complex and fluid shapes that mimic the variability of human handwriting. It provides a robust framework for the generation of synthetic handwritten text, which can be used for the training of OCR systems.

The only thing needed for such an approach is a collection of letter templates from a particular writer which can be gathered in a span of an hour, depending on the writing complexity. As we will demonstrate further handwriting can be varied automatically using augmentations during generation and post-processing stages so that we can imitate various aspects of handwriting. Though it can still be profitable to enlarge a template database with various ways of writing the same letter. On fig. 5 two variations of writing letter 'т' can be seen. During generation we can randomly pick template variations from different writers to create unique handwriting styles. During generation of a particular handwritten page we ensure that once style of each letter has been picked - it stays the same on the page, as it is highly unlikely for a real writer to spontaneously change writing style for a particular letter.

fig. 5 - two variations of letter 'т' in word 'этих'

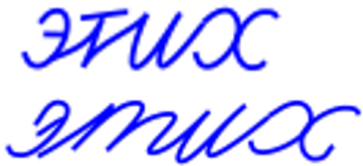

Information regarding letter template database collection with examples from a single writer is provided in Appendix A.

### 1.2.3 Minor generation techniques

There are some additional minor techniques we employ to further improve the quality of generated text. They are inconsistent with the general Bézier curve generation approach, so they will be described here.

**Diacritics imitation.** Some russian letters such as 'Й' and 'Ё' have parts that are similar to diacritic parts of letters in other languages. We describe these parts in a template using additional curves, so that augmentations are applied separately to the general and additional curve. Letter 'т' can also be written using two parts - lower one and the upper one, since writers that do so usually connect them in a particular way shown on fig. 6.

fig. 6 - Main and additional curves in 'ё' and 'й'

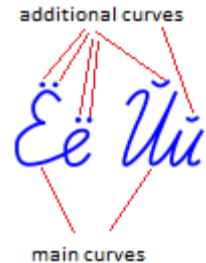

**Handwriting direction change.** Sometimes in cases of letters 'а', 'м' and 'л' instead of smooth transitions at part of the letters writers tend to write in a previously opposite direction. We imitate that using a special flag in a template that makes both vectors V in a single point have the same direction as shown on fig. 7.

fig. 7 - Direction change in letter 'а'

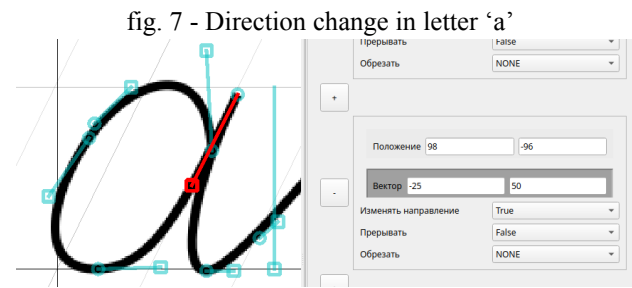

**Interruptions.** Some letters are usually written in a way that requires the writer to disconnect a pen from paper to continue writing. This is most evident in a case of the letter 'б' that has a hat on top. To imitate this behavior some points have a flag that prevents the next curve from generating, essentially finishing the letter.

**Cutoff.** In handwriting some letters end a bit differently when they are in the middle of a word and when they are located last. In the case of the letter 'л' when located on an end of a word there is a particular ending, which is incompatible when located in the middle of a word. Such letters possess a special flag that deletes the last letter point if it is in the middle of a word(see fig. 8).

fig. 8 - Letter cutoff demonstration

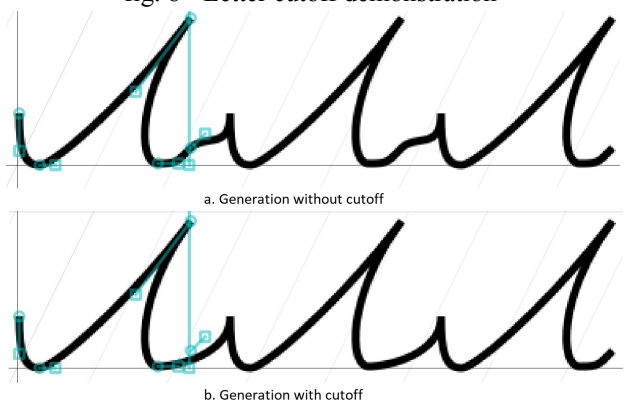

a. Generation without cutoff

b. Generation with cutoff

### 1.2.4 Handwriting augmentations

To further improve variability of handwriting we apply data augmentations at two stages:

**During handwriting style initialization.** It includes width of characters, distance between characters and between words and skew. These parameters are usually consistent for each writer, so we only vary them a bit using normal noise at the next stage. Mean value tends to stay the same.

**During curve generation.** We vary the above parameters in a small range near the initialized value. Additionally we add noise to each vector starting point location, vector rotation and length, letter location on Y-axis, writing speed and probability of character disconnection. In previously described diacritics imitation we additionally change the size of dots and caps.

In-depth review of all handwriting augmentations and parametrization with examples and explanations is provided in Appendix B.

## 1.3.  Datasets

In order to test POC model performance we need to gather corpora of HTR errors on the selected domain. To do so, we gather specific corpora that is close to the distribution that we anticipate for our HTR model. Then we continuously generate handwriting using this corpora and input it in the HTR model to receive recognition errors. These errors are aligned with initial text, used to generate handwriting in order to collect pairs of correct sentences and their counterparts with handwriting-specific mistakes. Resulting dataset is then used to fine-tune the Transformer model in Seq2Seq fashion for Post-OCR correction. Finally we evaluate performance of vanilla HTR and HTR-POC combination on real handwritten data.

Thus two types of datasets are needed for the training and evaluation, which will be described below.

### 1.3.1 Text corpora of handwritten essays

As mentioned in the introduction, our primary focus will be student's essays. To match such text distribution we scraped Russian Internet sites that host essay examples for students and practice books. Resulting corpora consists of a total of 3.91M sentences or 284M symbols.

For research purposes we provide gathered text data in our paper-related GitHub repository[9].

### 1.3.2 Synthetic handwritten dataset

For now code for generating handwritten text is still in development, but for evaluation purposes we provide a sample of generated handwritten essays that were used in the process of obtaining HTR errors. This dataset may be valuable for pre-training Cyrillic HTR models because of the scarcity of available HTR datasets. Sample may be found in abovementioned GitHub repository[9].

### 1.3.3 HTR(POC) performance evaluation datasets

In order to evaluate POC performance we chose 2 datasets in the domain of student handwritten essays that are publicly available and thus suitable for comparative research. We also evaluate metrics on our in-house student Essays Dataset.

### HWR200

The HWR200 dataset represents a comprehensive resource for the advancement of machine learning applications in the domain of handwritten text analysis. It consists of a collection of Russian handwritten text images, meticulously curated to facilitate a diverse range of research tasks, including but not limited to, character and handwriting recognition, visual question answering, near-duplicate detection, and text reuse identification.

The dataset encompasses a total of 30,030 images of handwritten texts, each rendered in Russian, spanning a storage size of 44 gigabytes. This substantial volume ensures a robust sample for training and validation purposes.

Text contributions are sourced from 200 distinct writers, ensuring a high degree of variability in handwriting styles. Such diversity is critical for developing models that are resilient to the idiosyncrasies present in real-world handwriting.

To simulate varying environmental conditions, each text image is captured in three distinct lighting scenarios: scanned, under poor lighting(BadPhotos), and under optimal lighting conditions(GoodPhotos). This tri-modal capture process enhances the dataset's utility in training models to be invariant to lighting conditions, a common challenge in text recognition tasks.

The dataset includes instances where different authors have transcribed identical texts, as well as 'reuses', where sentences are replicated across different text samples. This aspect allows for nuanced research into text consistency and variance across handwriting samples.

No additional preprocessing was required for preparing this dataset for evaluation purposes, as all images are cropped and full text ground truth(GT) data fully aligns with image contents.

### School_notebooks_RU

The School Notebooks Dataset constitutes a specialized corpus of images depicting school notebooks filled with handwritten notes in Russian. This dataset is designed to foster the development of advanced detection and Optical Character Recognition (OCR) models, providing comprehensive annotations that enable end-to-end text recognition from page images.

Comprising 1,857 images, the dataset presents a substantial variety of handwritten content, typically displaying two notebook pages per image. Through meticulous preprocessing, this initial set was expanded to 3,213 isolated page images, with each page individually cropped to facilitate precise OCR and text localization tasks.

Annotations within the dataset are meticulously structured in the COCO format[10], a common schema in computer vision that allows for detailed marking of object instances. The annotations encompass detailed descriptions for polygons outlining each handwritten word, enabling granular text localization at the word level. Additionally, each polygon is accompanied by a text translation attribute, enhancing the dataset's utility for multilingual OCR applications.

Given the dataset's COCO annotation format, additional preprocessing steps were imperative to align the full-text GT with the segmented word crops. This involved the careful cropping of individual pages and the subsequent alignment of GT annotations with these segmented images. Such preprocessing ensures that each word is distinctly identified, facilitating the training of models on word-level recognition tasks.

Moreover, the dataset features dual versions to account for the presence of teachers comments, which introduced complexity beyond the scope of our primary OCR objectives. To this end, two derivative datasets were prepared: one inclusive of teachers comments and another with these elements excluded, thereby allowing for a more focused approach to model evaluation.

The School Notebooks Dataset is particularly valuable for tasks that require an understanding of handwriting in educational contexts, presenting both the challenges of diverse handwriting styles and the varied quality of handwritten text in student notebooks. The segmentation of each word and the precise GT alignment underscores the dataset's suitability for developing models capable of word-level text detection and recognition, a crucial step in comprehensive handwritten text analysis.

### Student Essays dataset

The "Student Essays" dataset is an internally curated corpus designed to advance our Handwriting Detection and Recognition models. It is a substantial collection, instrumental for training and validating the performance of our algorithms in the recognition of handwritten essays.

The dataset comprises handwritten essays authored by students practicing writing essays. The respondents are instructed to write on a given topic with minimal corrections on standardized A4 paper. All essays are then scanned in a good quality in grayscale. This uniformity in presentation, alongside the specification of two distinct essay formats, provides a consistent framework for OCR analysis, which will be further elucidated in the dataset examples appendix of our paper.

To facilitate the development and benchmarking of our models, the dataset has been meticulously annotated for full-text recognition. This manual labeling process has been critical for obtaining precise metrics that reflect the true performance of our models. It is noteworthy that the models exhibit optimal performance on this in-house dataset, while datasets such as HWR200 and School Notebooks are treated as out-of-distribution data, offering a broader validation scope.

Remarkably, no additional preprocessing was necessary for this dataset, which underscores the clean and orderly nature of the source material. For the purpose of evaluating the Post-OCR Correction (POC) model's efficacy, an end-to-end (E2E) test subset was utilized. Importantly, to avoid any bias, the models were not trained on this evaluation data.

A unique aspect of our dataset is the methodical approach adopted to ensure a balanced representation of various handwriting styles within the evaluation set. By classifying handwriting into 33 distinct styles, we constructed a balanced sample from a larger dataset, thus allowing for a comprehensive assessment across a spectrum of handwriting characteristics.

The "Student Essays" dataset is composed of 520 distinct essays, each ranging from 1 to 6 pages, culminating in a total volume that is commensurate with the School Notebooks dataset. This scale ensures a robust platform for the development and fine-tuning of our models, as well as for the substantive analysis of their performance in real-world scenarios.

Data examples for each dataset are presented in Appendix C.

## 1.4. Text Detection, HTR and POC models

### 1.4.1 Text Detection model

For detecting handwritten text we employ Character Region Awareness for Text Detection (CRAFT) architecture[11]. The model embodies a novel convolutional neural network (CNN) architecture[12] that exhibits a profound capability for localizing text at the character level within natural scene images. It diverges from conventional text detection paradigms by emphasizing character regions as opposed to holistic word or text line entities. This granularity facilitates the detection of text with a high degree of precision, particularly in environments where text orientation, size, and spacing are varied and unpredictable.

CRAFT's architecture is particularly well-suited for detecting handwritten words due to its fundamental design that isolates characters as distinct entities. Handwriting, with its intrinsic variability in character shapes, sizes, and linkages, poses a challenge to traditional text detection systems that assume uniformity in text appearance. CRAFT's character-centric focus, combined with its affinity-based linking mechanism, allows it to adeptly handle the fluid and dynamic nature of handwritten text. The model can effectively identify and group characters into words, even when faced with the idiosyncratic alignments and spacings characteristic of handwritten script. The efficacy of CRAFT in these scenarios is a testament to its architectural ingenuity and its alignment with the complexities of handwritten text recognition.

The CRAFT model's nuanced approach to text detection aligns well with the heterogeneity of handwritten text, providing a robust solution for accurate localization and subsequent OCR processes.

### 1.4.2 HTR model

The Convolutional Recurrent Neural Network (CRNN)[13] framework represents a synergistic integration of convolutional and recurrent architectures, specifically tailored for the task of sequential data prediction inherent in handwriting recognition. This amalgamation harnesses the spatial hierarchy processing capabilities of Convolutional Neural Networks (CNNs) with the sequential data processing proficiencies of Recurrent Neural Networks (RNNs)[14], resulting in a potent model adept at deciphering the complexities of handwritten text.

The CRNN model is particularly suited for handwriting recognition due to several reasons:

**Flexibility in Handling Variability:** The variable nature of handwriting, with its diverse styles and unpredictable character spacing, necessitates a model capable of learning from irregular patterns. CRNNs excel in this aspect by learning to recognize and predict sequences without the need for pre-segmentation.

**Contextual Understanding:** The recurrent layer's ability to capture contextual dependencies is paramount in handwriting recognition, where the meaning and appearance of characters can be significantly influenced by adjoining characters.

**End-to-End Training:** The CRNN can be trained end-to-end, allowing for joint optimization of both feature representation and sequence prediction, which is beneficial for the complex task of handwriting recognition.

CRNN architecture stands as a paradigmatic model for automated handwriting recognition. Its design encapsulates the essential features required for interpreting the intricate and dynamic patterns of handwritten text, providing a robust and efficient mechanism for accurate text transcription in real-world applications.

### 1.4.3. POC model

The Post-OCR Correction model employed in this study is predicated upon the T5 (Text-to-Text Transfer Transformer) architecture[3], specifically the variant **'cointegrated/rut5-base-multitask'** available on HuggingFace[15]. This model represents a distilled adaptation of the larger **'google/mt5-base'**[16], retaining a subset of Russian and English embeddings, thereby optimizing the model's performance for bilingual text processing tasks.

Characteristics of the T5-based POC Model:

**Pretrained Language Model:** The foundation of the POC model is a pre-trained language model, which has been extensively trained on a diverse corpus encompassing both Russian and English languages. This pretraining endows the model with a nuanced understanding of language semantics, syntax, and structure.

**Fine-Tuning for Contextual Tasks:** Subsequent to pretraining, the model has undergone fine-tuning on various text-based tasks, such as translation, paraphrasing, text completion, text restoration, simplification, dialogue response generation, question answering, and headline generation. This multifaceted fine-tuning enhances the model's adaptability and its capability to understand and manipulate text at a granular level.

**Seq2Seq Framework:** In the context of Post-OCR Correction, the T5 model is harnessed in a sequence-to-sequence (Seq2Seq) fashion. It receives the full text output from the OCR model as input and is tasked with generating a corrected version of the text.

This paradigm effectively transforms OCR errors into a sequence prediction problem, where the model leverages its pretrained and fine-tuned knowledge to infer the intended text.

T5-based model advantages for Post-OCR Correction are following:

**Robust Error Correction:** Given its extensive pretraining and fine-tuning, the model is well-equipped to tackle the intricacies of OCR-generated text, discerning and rectifying errors that may be present, ranging from simple typographical errors to more complex contextual mistakes.

**Contextual Understanding:** The transformer-based architecture is particularly adept at understanding context, a critical factor in Post-OCR Correction where the meaning and correctness of a word or phrase can be highly dependent on its surrounding text.

**Flexibility and Adaptability:** The ability of the model to perform a variety of language tasks translates into a high degree of flexibility, allowing it to adapt its correction strategy based on the nature of the text it encounters.

In this research, the 'cointegrated/rut5-base-multitask' model's unique characteristics are leveraged to directly address the challenges posed by OCR errors in Cyrillic text. By utilizing a T5 transformer pre-trained on a relevant language domain and fine-tuned on tasks aligned with text correction, the model is well-positioned to provide accurate Post-OCR corrections. This approach not only refines the text for improved readability and utility but also contributes to the broader goal of enhancing digital text quality in multilingual OCR applications.

Text Detection and HTR model training specifics are beyond the scope of this paper. However, we delve into the particulars of fine-tuning the T5 model for Post-OCR Correction (POC) tasks. The model was fine-tuned using the 'cointegrated/rut5-base-multitask' checkpoint as a starting point. For the optimization process, we employed the Adam optimizer[17], widely recognized for its efficiency in handling sparse gradients on noisy problems. The learning rate was set to a constant 1e-5, a value that balances the need for fast convergence with the risk of overshooting the minimum loss.

The loss function used was CrossEntropy, a standard choice for classification tasks, which in our case is apt for the sequence prediction required in POC tasks. The training involved a substantial dataset of 836,000 sample sentences, each containing OCR errors. This large dataset ensures comprehensive exposure to a variety of error patterns, crucial for the model's ability to generalize well to unseen data.

Batch processing was implemented with a size of 32, optimizing the balance between computational efficiency and memory constraints. The training was conducted over 12 epochs, which was the point at which the model began showing signs of overfitting. The entire training process spanned one day on a single Nvidia 2080 Ti GPU, demonstrating the model's efficiency in learning from large datasets within a reasonable timeframe. Additionally, a dropout rate of 0.1 was utilized to prevent overfitting, adding an element of regularization that helps in improving the model's generalization capabilities.

## 2. Results

In this part we present evaluation results for each dataset.

In the evaluation of our handwriting recognition models, the JiWER library[18] is employed to calculate key performance metrics. JiWER is a python package originally developed for assessing automatic speech recognition systems, but its utility extends to any domain requiring measurement of transcription accuracy. It provides a suite of metrics that are essential for gauging the effectiveness of our models, including Word Error Rate (WER), Match Error Rate (MER), Word Information Lost (WIL), Word Information Preserved (WIP), and Character Error Rate (CER).

These metrics are derived from the minimum-edit distance algorithm implemented in RapidFuzz—a high-performance library[19] that leverages C++ for computational efficiency. By computing the minimum number of insertions, deletions, and substitutions required to transform the hypothesis text output by our models into the reference text, we obtain a quantitative measure of model accuracy.

To ensure a comprehensive analysis, metrics are computed under three distinct preprocessing conditions: raw text without preprocessing(Raw text), text normalized to lowercase(Lowercase only), and text stripped of all non-alphabetic symbols(Only alphabetical). This tripartite evaluation strategy allows us to understand model performance across different levels of textual normalization, providing insights into the robustness of our handwriting recognition models under varying input conditions.

### 2.1. HWR200

The WAR and CAR evaluation metrics for HWR200 dataset are presented in Tables 1-2 respectively. Each domain is calculated separately for better comparison of medium impact on recognition and correction quality. Correction example can be seen in fig. 9.

**Scans.** The application of POC consistently improves the mean WAR across all normalization types. The results indicate that different normalization strategies impact the accuracy rates. Specifically, the Only alphabetical approach, which retains only alphabetical characters, yields the highest mean WAR and CAR, both with and without POC. This is due to the fact that removing non-alphabetic characters simplifies the text structure. The improvement in mean CAR with POC is less pronounced compared to WAR. This could imply that the model's character-level recognition capabilities are already quite high, and the POC model offers incremental improvements in this aspect. The relatively high baseline CAR values even without POC may be attributed to the fact that the Text Detection and Text Recognition models were trained on scans of handwritten student text. This training likely tailored the model to perform well on similar datasets, such as the one evaluated here.

**GoodPhotos.** Both the mean WAR and CAR are notably lower on photos compared to scans. This trend is consistent across all normalization types and suggests that the OCR and TD models are less effective in interpreting text from photos, even under good lighting conditions. While the application of POC improves accuracy in both scenarios, the degree of improvement is more modest in photos. This indicates that the POC model, though beneficial, is less adept at compensating for the challenges presented in photo-based text recognition. An interesting observation is the relatively high baseline CAR on photos without POC, especially when compared to the lower WAR. This suggests that while word-level recognition is significantly more challenging in photos, character-level recognition remains relatively stable.

**BadPhotos.** There is a notable decrease in accuracy for photos in bad lighting compared to good lighting and scans. This stark difference underscores the critical role of lighting quality in OCR performance. While POC still improves accuracy in bad lighting conditions, its effectiveness is substantially reduced compared to scans and good lighting photos. This suggests that the POC model, despite its capabilities, struggles to fully compensate for the challenges introduced by poor lighting, such as increased noise and decreased text clarity. Interestingly, the baseline CAR without POC is relatively high even in poor lighting, similar to the results in good lighting and scans. This suggests that character-level recognition is less sensitive to lighting conditions compared to word-level recognition.

When comparing all three conditions (scans, good lighting photos, bad lighting photos), it's evident that the quality and clarity of the source medium play a significant role in OCR accuracy. Scans, offering the highest quality and consistency, lead to the best performance, followed by photos in good lighting.

Photos in poor lighting present the most challenging scenario for both the OCR/TD and POC models.

fig. 9 - Correction example in HWR200 dataset

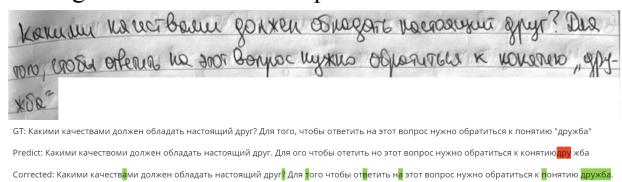

Table 1: WAR metric on HWR200 dataset

| Normalization | WAR without POC, % | WAR with POC, % |
|---|---|---|
| **Scans** | | |
| Raw text | 43.245 | **56.045** |
| Lowercase only | 48.468 | **59.381** |
| Only alphabetical | 57.130 | **66.234** |
| **GoodPhotos** | | |
| Raw text | 27.722 | **39.724** |
| Lowercase only | 31.811 | **42.809** |
| Only alphabetical | 38.846 | **49.043** |
| **BadPhotos** | | |
| Raw text | 16.132 | **24.810** |
| Lowercase only | 19.007 | **27.089** |
| Only alphabetical | 24.273 | **32.030** |

Table 2: CAR metric on HWR200 dataset

| Normalization | CAR without POC, % | CAR with POC, % |
|---|---|---|
| **Scans** | | |
| Raw text | 78.713 | **78.772** |
| Lowercase only | **80.027** | 79.552 |
| Only alphabetical | **81.717** | 80.781 |
| **GoodPhotos** | | |
| Raw text | **68.683** | 67.347 |
| Lowercase only | **69.932** | 68.170 |
| Only alphabetical | **71.702** | 69.447 |
| **BadPhotos** | | |
| Raw text | **50.908** | 49.537 |
| Lowercase only | **52.114** | 50.270 |
| Only alphabetical | **53.563** | 51.318 |

## 2.2 School_notebooks_RU

Metrics for School_notebooks_RU dataset are presented in Tables 3-4. We evaluate performance for both versions of the dataset with and without Teacher comments. Correction example for this dataset can be seen in fig. 10.

**Student-only text.** When POC is applied, there is an improvement in the mean WAR for student-only handwriting across all normalization types. This indicates that POC effectively enhances word-level recognition in student handwriting. Conversely, the mean CAR decreases with the application of POC. This suggests that while POC improves the recognition of word sequences, it might introduce inaccuracies at the character level, possibly due to over-correction or misalignment in the context of student handwriting. The improvement in WAR coupled with the reduction in CAR implies that the POC model is better tuned for enhancing the overall structure and flow of words in a sentence, but this comes at the expense of individual character accuracy.

**Student text with Teacher comments.** Including teacher comments in the GT leads to a decrease in accuracy for both mean WAR and CAR compared to the student-only handwriting version. This decrease is observed both with and without the application of POC. With POC, there is an increase in mean WAR compared to the results without POC, however, the mean CAR decreases with POC. This pattern suggests that while POC improves word-level recognition when teacher comments are included, it adversely affects character-level accuracy. Comparing this version of the dataset with the student-only version reveals that the inclusion of teacher comments poses additional challenges for the OCR models. The accuracy rates are lower in this version, suggesting that the complexity introduced by teacher comments impacts the model's ability to recognize and correct text accurately. The decrease in CAR with POC, especially in the version with teacher comments, indicates that the POC model may struggle to adapt to more complex datasets containing varied handwriting styles. This is likely due to the increased complexity and variability introduced by the mix of student and teacher handwriting.

fig. 10 - Correction example in School_notebooks_RU dataset

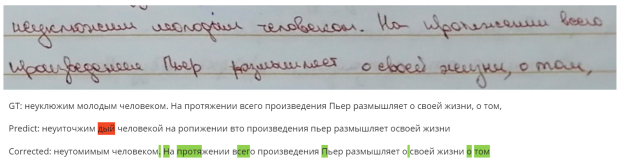

Table 3: WAR metric on School_notebooks_RU dataset

| Normalization | WAR without POC, % | WAR with POC, % |
|---|---|---|
| **Student text** | | |
| Raw text | 20.822 | **24.351** |
| Lowercase only | 25.851 | **28.590** |
| Only alphabetical | 32.634 | **35.223** |
| **Student text with Teacher comments** | | |
| Raw text | 17.770 | **21.557** |
| Lowercase only | 23.104 | **26.025** |
| Only alphabetical | 30.749 | **33.450** |

Table 4: CAR metric on School_notebooks_RU dataset

| Normalization | CAR without POC, % | CAR with POC, % |
|---|---|---|
| **Student text** | | |
| Raw text | **67.114** | 62.879 |
| Lowercase only | **68.854** | 64.274 |
| Only alphabetical | **71.684** | 66.793 |
| **Student text with Teacher comments** | | |
| Raw text | **67.650** | 63.364 |
| Lowercase only | **69.413** | 64.775 |
| Only alphabetical | **71.987** | 67.106 |

## 2.3 Student Essays

Metrics results for this dataset are demonstrated in Tables 5-6. Correction example for this dataset can be seen in fig. 11.

The application of POC leads to a notable improvement in WAR across all normalization types. This substantial improvement suggests that POC is highly effective in enhancing word-level recognition in high-quality scan environments. Unlike in the previous datasets, the POC also improves the CAR, though the increase is more modest than in WAR. This improvement in CAR, coupled with the significant gains in WAR, indicates that the POC model is well-suited to the high-quality nature of this dataset. The base OCR model demonstrates a strong performance in both WAR and CAR before POC is applied, especially when compared to the other datasets. This is due to the high quality of the scans and the fact that the TD and OCR models were trained on data from the same distribution. The in-house Dbrain dataset's results highlight the optimal conditions under which the POC model excels. In a scenario with high-quality scans and minimal extraneous markings, the POC model significantly enhances both word and character recognition accuracy. Another possible explanation is that in the process of gathering HTR errors on generated handwritten data we did not account for lighting conditions to simulate errors caused by this phenomenon. This explains why POC model excels on in-house dataset and shows adequate results on HWR200-Scans dataset, while demonstrating poor performance on data that had photos or bad lighting conditions.

fig. 11 - Correction example in Student Essays dataset

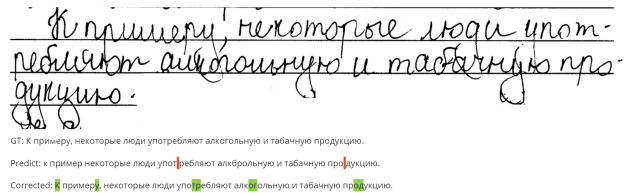

Table 5: WAR metric on Student Essays dataset

| Normalization | WAR without POC, % | WAR with POC, % |
|---|---|---|
| Raw text | 65.425 | **77.452** |
| Lowercase only | 71.418 | **80.079** |
| Only alphabetical | 78.173 | **85.384** |

Table 6: CAR metric on Student Essays dataset

| Normalization | CAR without POC, % | CAR with POC, % |
|---|---|---|
| Raw text | 91.305 | **92.259** |
| Lowercase only | 92.501 | **92.776** |
| Only alphabetical | **93.683** | 93.619 |

## 3.  Discussion

The results and analysis from our study on the Post-OCR Correction (POC) model, in conjunction with Text Detection (TD) and HTR models, reveal several key insights and implications for the field of handwritten text digitization and correction.

Firstly, the POC model demonstrates notable proficiency in enhancing both WAR and CAR metrics, particularly when the TD and OCR models are trained on datasets that closely align with the distribution and handwriting generation parameters used for creating POC error generation. This alignment underscores the importance of dataset consistency across the training pipeline to achieve optimal OCR performance.

However, our approach exhibits diminished effectiveness on out-of-distribution data. Despite this, there is still a discernible increase in WAR in such scenarios. This improvement can be attributed to the POC model's ability to leverage the contextual understanding of written sentences, allowing it to infer correct words based on their meaning within the sentence structure.

To bridge the gap in performance on out-of-distribution data, a potential strategy involves integrating some of this diverse data into the training process for the TD and OCR models. Additionally, simulating various photographic conditions when generating data for POC training might contribute to enhancing the model's adaptability and accuracy across different data types.

The performance on the HWR200 scans further supports the theory that our model combination is particularly well-suited for high-quality scans, as these conditions closely resemble those of our in-house dataset. This finding suggests that the model's effectiveness is partly contingent on the quality and characteristics of the input data.

A consistent observation across all datasets is the improvement in WAR metrics with the application of POC, validating our approach of using realistically generated handwriting with Bézier curves to create large error corpora for POC training. This approach successfully enhances the model's capability to correct word-level errors.

To address the observed decrease in CAR while preserving the gains in WAR, we propose implementing techniques that modulate the POC model's application. For instance, utilizing a confidence threshold to selectively correct only those letters or words with low OCR/TD confidence could mitigate the drop in character accuracy while maintaining WAR gains.

Finally, to specifically enhance performance on the School_notebooks_RU dataset, integrating a text crop classification model into our pipeline could be beneficial. By classifying text crops as student handwriting, markings, or teacher handwriting, we can create separate End-to-End datasets for each category. This targeted approach would allow for more precise model training and reduce the likelihood of errors, particularly in datasets characterized by a mix of different handwriting styles and annotations.

In conclusion, our study highlights both the strengths and areas for improvement in the current state of POC technology for handwritten text. The insights gained point towards a future where more sophisticated and context-aware OCR systems can be developed, offering enhanced accuracy and adaptability across a wider range of handwritten text scenarios.

## 4.  Related work

In the realm of OCR error correction, several notable studies provide valuable insights and methodologies relevant to our research. The paper "Optimizing the Neural Network Training for OCR Error Correction of Historical Hebrew Texts"[20] addresses the challenge of OCR errors in historical documents. It proposes an innovative method for training neural networks using less manually created data, focusing on generating language and task-specific training data. Their findings emphasize the effectiveness of this approach over randomly generated errors and the importance of dataset genre and area in training neural networks for OCR post-correction.

Another significant contribution is from "Lights, Camera, Action! A Framework to Improve NLP Accuracy over OCR documents"[21] which tackles OCR errors impacting downstream NLP tasks. They develop a document synthesis pipeline to generate realistically degraded data, demonstrating that a text restoration model trained on this data significantly mitigates OCR errors, even in out-of-domain datasets. The POC model used is based on a novel approach that we call Levenshtein-based model, as for each token in input sequence it predicts an action such as 'insert', 'delete', 'pass' or 'replace' and a symbol from vocabulary to be used with this action.

The study "Lexically-Aware Semi-Supervised Learning for OCR Post-Correction"[22] explores neural post-correction methods for less-well-resourced languages. They introduce a semi-supervised learning method utilizing raw images for performance improvement and a lexically-aware decoding method augmenting neural models with a count-based language model. This approach achieves significant error reductions, demonstrating the value of self-training and lexically-aware decoding.

In "Post-OCR Document Correction with large Ensembles of Character Sequence-to-Sequence Models"[23] a novel method is proposed for extending sequence-to-sequence models to process longer sequences efficiently. Applied to OCR text correction in multiple languages, their method involves splitting documents into character n-grams and combining corrections using a voting scheme, achieving state-of-the-art performance in several languages.

Lastly, "Leveraging Text Repetitions and Denoising Autoencoders in OCR Post-correction"[24] presents a model that estimates OCR errors from repeating text spans in large OCR text corpora. They generate synthetic training examples based on this error distribution, using them to train a character-level neural seq2seq model. Their results show significant improvements over existing OCR systems and models utilizing uniformly generated noise.

Each of these studies contributes unique methods and insights, ranging from specialized training data generation, semi-supervised learning approaches, ensemble model strategies, to innovative error estimation techniques, all of which are instrumental in advancing OCR post-correction technology.

## 5. Further work

In our future work, we plan to enhance the performance and versatility of the POC model in several ways. Firstly, we intend to introduce photographic simulation into the handwriting generation process for POC training. This addition aims to better replicate the diverse conditions encountered in real-world scenarios, thereby improving the model's adaptability and accuracy across various photographic environments.

Another area of focus will be on refining the application of the POC model by restricting its usage to instances where OCR or TD models exhibit low confidence. This targeted approach is expected to optimize the balance between improving WAR and maintaining CAR, thereby enhancing overall accuracy without introducing new errors.

Expanding our research to include multiple languages will also be a key direction. By applying our approach to other languages with distinct handwriting characteristics, we can assess the universality and scalability of our methodology, potentially leading to broader applications in global OCR tasks.

Furthermore, we plan to evaluate our approach using different architectural frameworks. This includes experimenting with RNNs, Levenshtein operations-based POC models, and other transformer models such as BERT[25] and GPT[26]. These explorations aim to identify alternative or complementary architectures that may offer advantages in specific aspects of OCR and POC tasks.

Through these initiatives, we aspire to advance the field of OCR and POC, making it more robust, adaptable, and applicable across a wider range of languages and scenarios.

## Appendix A. Character template database

Here we provide a whole set of all available letters in the Russian alphabet collected from a single writer as can be seen in fig. 12.

fig. 12 - Russian Character template database example

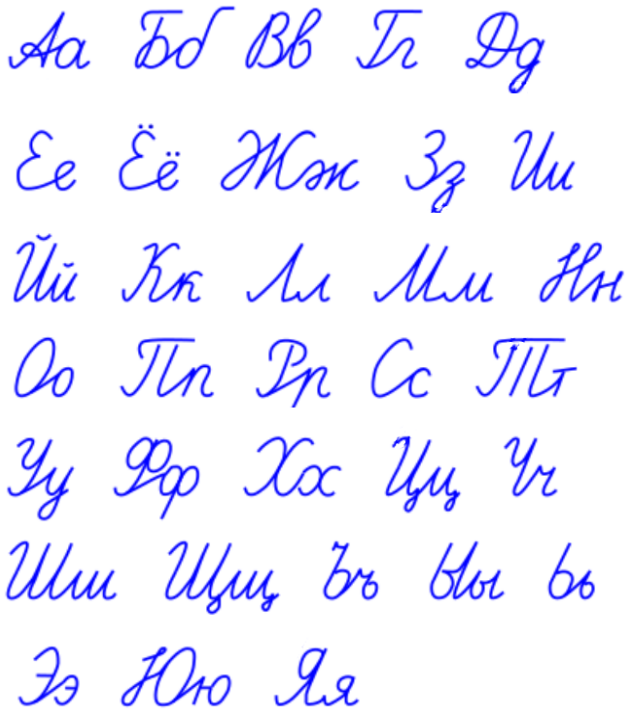

Collection process is straightforward. We locate the desired style of handwriting and reproduce it by manually inputting control points for each letter on a canvas. The process is then repeated for each letter in an alphabet and lowercase letters. Our observations show that it takes approximately 1 hour to collect the full alphabet from a single writer.

For ease of use we also implement transparent layers, so that letters can be copied pixel-perfect with an immediate reference. This way we created first version of Russian character database by simply copying

## Appendix B. Handwriting curve generation augmentations.

As we described earlier in 1.2.4. We use two types of augmentations. First type is initialized for a whole page, while the second one may change even across a single word. Here we provide demonstration of all controllable values in three states:

- No value inputted, thus no augmentation of that kind is used
- Medium value augmentation
- Intense augmentation

Note that the latter usually leads to unrecognizable text, which is unfavorable.

**Point-noise augmentation.** This augmentation (fig. 13) refers to the noise added during the rendering of a point in the process of generating handwritten text using Bézier curves. In handwriting, no two letters are written exactly alike, and this augmentation reflects that natural variability. When a letter's point is being drawn, an offset is applied to it. At a zero noise level, all letters would appear uniform, resembling characters in a handwriting practice book. As the 'point_noise' value increases, the letters start to appear more natural and exhibit slight variations, akin to actual handwriting. However, excessively high values of this parameter can lead to the letters becoming overly distorted and unrecognizable, resulting in a visually chaotic appearance. This augmentation thus plays a critical role in simulating the realistic variations found in human handwriting.

fig. 13 - noise-p augmentation

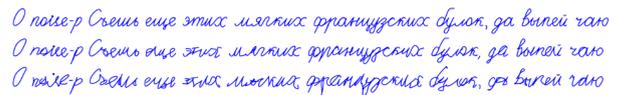

**Noise vector length and Noise vector Rotation augmentations.** These augmentations are applied to the vector components of Bézier curves, which consist of sets of points and vectors.

Vector Rotation Noise(fig. 14) modifies the direction of the vectors associated with each point of the Bézier curve. Adjusting the direction of these vectors can alter the shape of the curve, thereby impacting the structure of the generated handwriting. It's important to apply this augmentation cautiously, as even minor changes in the vector direction can significantly affect the visual appearance of the letters.

Vector Length Noise(fig. 15) involves changing the length of the vectors in the Bézier curves. Altering the vector length affects how far the curve extends from each point, thus modifying the overall look of the handwritten text. Similar to vector rotation noise, changes in vector length should be applied with restraint. Excessive alterations can lead to disproportionate and unnatural-looking letters.

Both these augmentations require careful calibration. Even slight changes in the vectors that form the Bézier curves can substantially alter the appearance of the handwriting, potentially making it look unnatural or rendering it illegible. Therefore, these augmentations should be employed with minimal values to ensure that the generated handwriting maintains a realistic and coherent form.

fig. 14 - Vector Rotation Noise augmentation

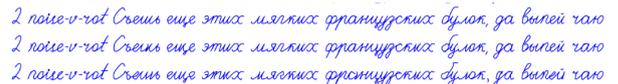

fig. 15 - Vector Length Noise augmentation

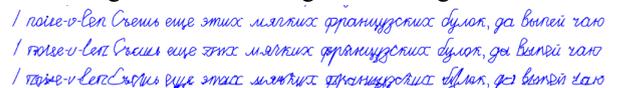

**Skew augmentation.** The "skew" augmentation(fig. 16) refers to the modification of the slant or tilt in the generated handwriting. In natural handwriting, there is typically a slight rightward slant. However, this slant can vary considerably among different individuals. Some people write with nearly vertical strokes, which can sometimes even lean to the left, while others exhibit a more pronounced rightward slant, with their letters appearing as if they are 'lying' to the right.

This augmentation allows for the simulation of these variations in slant, thereby enhancing the realism of the generated handwriting. By adjusting the skew of the letters, the handwriting can be made to resemble the range of natural inclinations found in human writing. This is particularly important for creating a diverse dataset that captures the idiosyncrasies of individual handwriting styles, allowing for more effective training and evaluation of handwriting recognition models. Careful application of this augmentation is crucial to ensure that the generated handwriting maintains a natural and legible appearance.

fig. 16 - skew augmentation

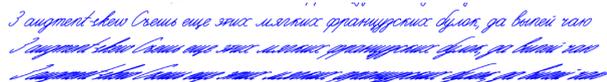

**Character Distance and Space Distance Standard Deviation augmentation.** Address the variability in spacing found in natural handwriting, specifically between characters and words, respectively.

Character Distance Standard Deviation augmentation(fig. 17) varies the distance between individual letters. In natural handwriting, the spacing between letters is not uniform but varies slightly. To simulate this, a random additive factor, delta, is introduced to the standard spacing between characters. This addition is carefully regulated, especially on the left side, to prevent the subsequent letter from overlapping excessively with the previous one. This controlled variability in character spacing enhances the realism of the generated handwriting, reflecting the subtle irregularities inherent in human writing.

Space Distance Standard Deviation(fig. 18) is similar to the character distance augmentation, but this one focuses on the spacing between words. The spacing between words in natural handwriting also varies and is not consistently the same. By applying a random additive factor to the standard space between words, this augmentation replicates the natural variations seen in word spacing. As with character spacing, careful application is crucial to maintain the legibility and natural appearance of the handwriting.

Both augmentations play a crucial role in generating handwriting that closely mimics the natural variations and irregularities of human writing. By introducing controlled randomness in the spacing of characters and words, these augmentations contribute to creating a diverse and realistic dataset, which is vital for the effective training and evaluation of handwriting recognition models.

fig. 17 - char-distance-std augmentation

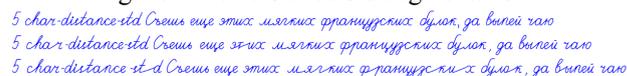

fig. 18 - space-distance-std augmentation

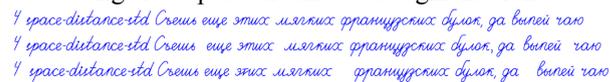

**Point-size Standard Deviation augmentation.** The augmentation(fig. 19) specifically pertains to the size of the points in the handwriting, relevant to letters such as 'Ё' and punctuation marks like the period. This augmentation is designed to adjust the size of the additional curves in these letters and any diacritical marks.

In natural handwriting, the size of such points and diacritical elements can vary, contributing to the unique characteristics of individual writing styles. The "point_size_std" augmentation introduces variability in these elements by applying a standard deviation to their sizes. This adjustment ensures that the points and diacritical marks do not appear uniformly sized across different instances, thereby enhancing the realism of the generated handwriting.

This augmentation is particularly significant for languages that include letters with additional curves or diacritical marks, as it allows the handwriting generation process to faithfully replicate the subtle variations in point sizes that occur naturally. By incorporating this level of detail, the generated handwriting becomes more representative of actual human writing, improving the diversity and authenticity of the handwriting dataset used for training and evaluating recognition models.

fig. 19 - point-size-std augmentation

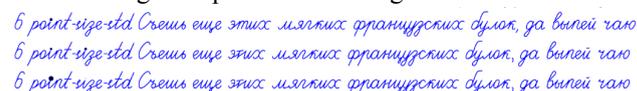

**Width augmentation.** This augmentation(fig. 20) relates to modifying the width of letters in the generated handwriting. In natural handwriting, individuals exhibit variations in how wide or narrow they write their letters. Some people may write with broader strokes, resulting in wider characters, while others might have a tendency towards more slender, narrow letter forms.

fig. 20 - width augmentation

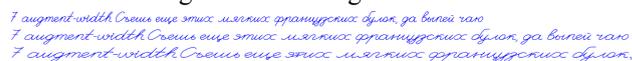

**Y-axis augmentation.** The "y_delta_max" and "y_delta_speed" augmentations are designed to simulate the natural phenomenon of the 'dancing line' in handwriting. Writing a long line perfectly straight is nearly impossible; it tends to deviate slightly upwards or downwards. However, when writing on lined paper, the extent of this 'dancing' or deviation is typically less pronounced.

y_delta_max parameter(fig. 21) controls the maximum extent to which the letters can deviate from an ideal horizontal line, either upwards or downwards. It sets the limit for how far the line of text can 'wander' from the baseline.

y_delta_speed parameter(fig. 22) determines the frequency and rate at which the line of text shifts above or below the ideal line. It essentially governs how quickly and often the text 'drifts' in its vertical position.

Instead of employing simple clipping to keep the deviations within bounds, a 'soft' clipping method using the hyperbolic tangent function is used. This approach ensures that the ease of moving up or down decreases as the deviation approaches the threshold value. This method prevents situations where the line of text abruptly shifts from moving upwards to sharply going downwards, maintaining a more natural flow in the vertical movement of the text.

By incorporating these augmentations, the generated handwriting can mimic the subtle vertical fluctuations that are characteristic of natural handwriting.

fig. 21 - y-delta-max augmentation

fig. 22 - y-delta-speed augmentation

**Symbol Disconnection augmentation.** This augmentation(fig. 23) addresses the variability in how individuals connect or disconnect letters within a word in handwriting. In reality, people's practices in connecting letters in a word vary significantly. Some consistently connect the letters, while others frequently leave them disconnected.

This augmentation introduces a probability parameter that determines the likelihood of letters within a word being disconnected. By setting this parameter, it's possible to simulate both writing styles: one where letters are typically joined and another where they are more often separated.

fig. 23 - symbol-disconnect-prob augmentation

**Appendix C. Handwritten dataset examples.**

Here we provide image examples for each dataset with a description.

Students Essays dataset is presented in two variations on fig.24 and fig.25. Note that the first variation has lines for increased writing consistency, improving recognition quality. Second variation poses more challenges as usually handwriting tends to have a tilted line of writing.

fig. 24 - First page type in Student Essays dataset

fig. 25 - Second page type in Student Essays dataset

HWR200 dataset image sample is presented on fig. 26, specifically sample from Scanned part of dataset. It bears a similar trait to the first page type of Student Essays dataset, but the images are usually of a bit lower scan quality.

fig. 26 - Scanned page example for HWR200 dataset

School_notebooks_RU dataset sample is presented on fig. 27. Image quality is even lower than in the HWR200 dataset and all images are photos instead of scans. Teacher comments are usually in red pen color, while student handwriting is in blue pen color.

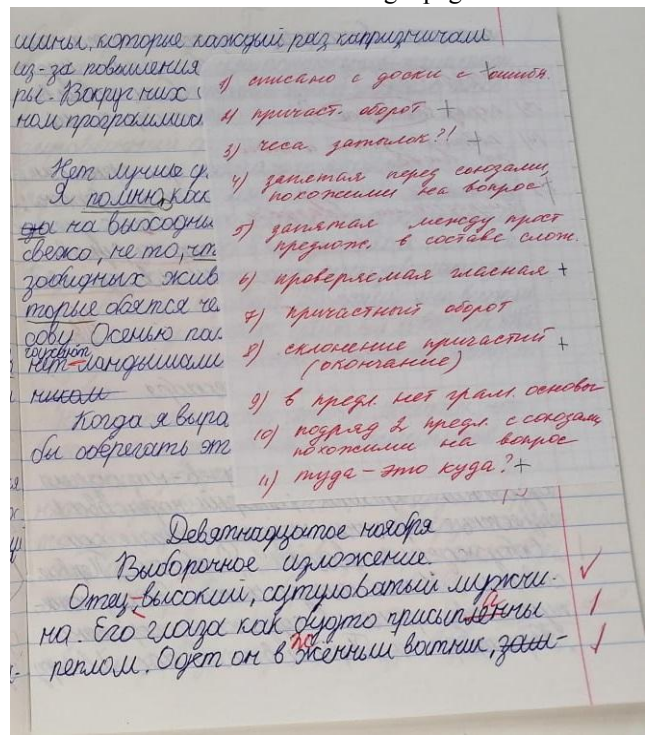

fig. 27 - Page example for School_notebooks_RU dataset. Student handwriting(blue) and Teacher comments(red) can be seen within a single page